\documentclass{llncs}

\usepackage{graphicx}
\usepackage{llncsdoc}

\usepackage{todonotes}

\begin{document}
\title{Highly Automated Learning for Improved \\ Active Safety of Vulnerable Road Users}
\author{Maarten Bieshaar\inst{1} \and G\"unther Reitberger\inst{2} \and Viktor Kre\ss{}\inst{3} \and Stefan Zernetsch\inst{3} \and
	Konrad Doll\inst{3} \and Erich Fuchs\inst{2} \and Bernhard Sick\inst{1}}
\institute{Intelligent Embedded Systems Lab, University of Kassel, Germany, \\
	\email{\{mbieshaar, bsick\}@uni-kassel.de}
	\and FORWISS, University of Passau, Germany, \\
	\email{\{reitberg, fuchse\}@forwiss.uni-passau.de}
	\and KoopAutoV, University of Applied Sciences Aschaffenburg, Germany,\\
	\email{\{viktor.kress, stefan.zernetsch, konrad.doll\}@h-ab.de}}

\maketitle

\begin{abstract}
	
Highly automated driving requires precise models of traffic participants. 
Many state of the art models are currently based on machine learning techniques.
Among others, the required amount of labeled data is one major challenge. 
An autonomous learning process addressing this problem is proposed. 
The initial models are iteratively refined in three steps: 
(1) detection and context identification, (2) novelty detection and active learning and (3) online model adaption.
	
\end{abstract}


\section{Introduction}\label{sec:Introduction}

A basic prerequisite for highly automated driving in urban areas are precise models of the 
surrounding (traffic) environment of the vehicles. These models are able to detect intentions of other road users and forecast their future trajectories such that safe driving strategies can be 
realized. In future traffic scenarios, vehicles and other road users communicate, exchange information, and cooperate on various levels~\cite{BRZ+17}. 
Vulnerable road users (VRU), e.g. pedestrians, cyclists etc., are important participants in today's and future traffic.
The behavior of VRUs is highly dynamic, situation- and context-dependent and often not in compliance with traffic regulations.
Moreover, there are a variety of different VRU classes, e.g. pedestrians, cyclists, and skateboarders, which all behave differently.
For models based on machine learning techniques, a considerable amount of sample data 
is required. This data has to be acquired in time-consuming and costly experiments, requiring high precision sensors which are used as 
reference data. 

In this article we propose a concept which shall allow to gather sample data using close to serial production sensors to efficiently create and enhance detection and prediction models, i.e. alleviating the necessity to perform costly experiments 
and minimize the required amount of human interaction (e.g labelling effort).


\section{Architecture description}\label{sec:Architecture}

\begin{figure}[htbp]
	\centering
	\includegraphics[scale=0.25]{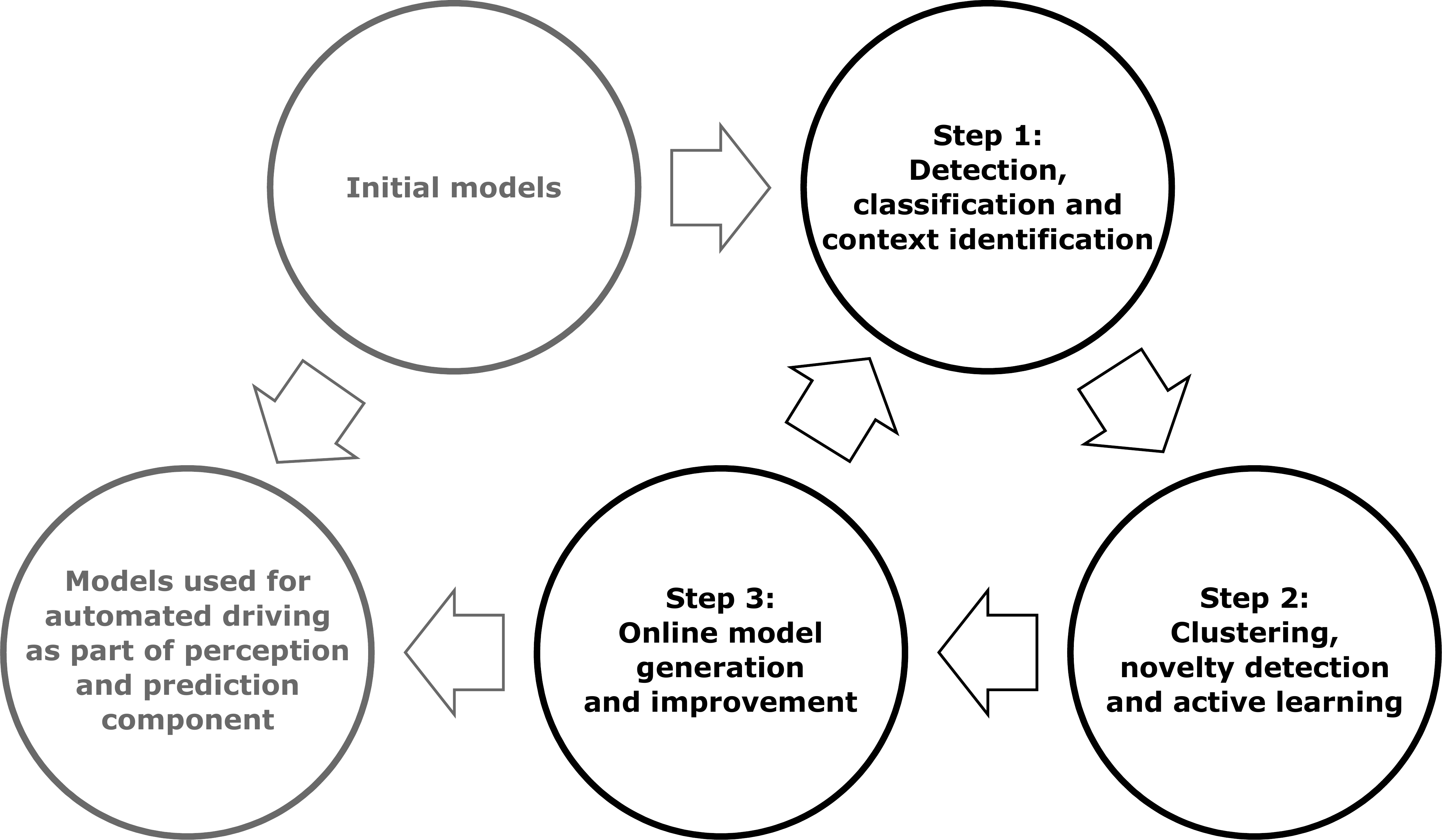}
	\caption{Highly automated learning process.}
	\label{archtitecture}
	\vspace{-10mm}
\end{figure}

We propose a holistic approach consisting of three fundamental stages for continuous autonomous learning as described in 
Figure~\ref{archtitecture}.
Sample data is gathered in a goal-oriented fashion in the everyday traffic using close to serial production sensors of vehicles, e.g. camera or radar. In particular, no high precision sensors are used.
The acquired data is used to constantly improve existing detection and prediction models, to integrate context information, 
e.g. time of day,  traffic volume and location information, or to automatically create new models, 
e.g. for new or so far unconsidered VRU classes.
Detection, classification, and context identification are performed in the first step.
To cover wide ranges of different VRU classes in daily traffic, e.g. pedestrians with strollers or skateboarders, 
the gathered data is autonomously clustered and divided into new classes.
The necessity of feedback from human experts shall be minimized.
This requires novelty detection and active learning which are implemented in the second step. 
Finally, in the third step motion prediction and detection models are generated and improved based on the clustered data. 
These models are used for automated driving as part of perception and prediction component. Moreover, these models also improve the detection and classification in the first step. To bootstrap the iterative learning process, initial models based on labeled data are used.
\vspace*{-0.35cm}

\section{Detection, classification, and context identification}\label{sec:GroundTruth}

The first important challenge in this context forms the generation of a "ground truth". 
This is usually time consuming or based on expensive equipment in current processes, 
because the positions of the detected traffic participants have to be determined as precisely as possible. 
We aim to achieve a precise self-localization with close to serial production sensor systems and high-accuracy maps~\cite{Sch+13}. 
The detection and classification of VRUs~\cite{KGB+13} should, as well, 
be based on production sensor systems such as stereo cameras, laser scanners or radar.
The detection is extended by tracking~\cite{MRW+13} based on continuously refined motion models.
As soon as a map of the environment is given, we intend to combine the tracking step with a semantic classification of the situation, 
e.g. we aim to determine if the pedestrian is walking in the street or on the sidewalk.

Since a real ground truth is not available with the equipment intended and in order to be able to evaluate the results, 
we aim to develop quality and certainty measures concerning the detection and position of a VRU. 

\section{Clustering, novelty detection, and active learning}\label{sec:Learning}

The second step is concerned with the assessment of classification and detection results of the former step with respect to the models used for detection and prediction. Prediction and detection models have to assess themselves, i.e. to decide autonomously whether a particular observed behavior of a VRU can be used to improve prediction quality (e.g. detection of unknown and novel VRU classes or unusual trajectories). 
Innovative active learning~\cite{CLL+16,Set10}, novelty detection~\cite{GS16}, and clustering techniques allow the system 
to select only valuable information such that it can cope with a massive amount of data.
Moreover, semi-supervised learning methods enable the system to make use of structural and non-labelled information~\cite{RCS15}, thus supporting the active learning process. 
Using these techniques, the system is capable to autonomously decide from which data source and sample it can extract knowledge and henceforth it is able to learn efficiently.


\section{Online model generation and improvement}\label{sec:Model}

In the third step we develop algorithms to improve existing models to predict the future trajectories of traffic participants 
\cite{GKD+15,ZKG+16} at runtime through self-learning techniques. 
To further improve the prediction we not only distinguish between different classes of traffic participants, 
but we also subdivide the classes based on their behavior in different situations, 
e.g. pedestrian on the sidewalk or pedestrian crossing the street.
Compared to data generated in specific series of experiments, the available data is less accurate, 
but it consists of a greater number of examples, 
which are more realistic and cover behavior patterns that cannot be created in an experiment, e.g. VRU behavior which is not 
in compliance with the road safety regulations.

Furthermore, we investigate how additional knowledge, such as the time of day or the traffic light phase can be used to improve the prediction process.

\section{Discussion}\label{sec:Conclusion}

In this article we presented a concept to cost-efficiently improve VRU detection and intention prediction models based on machine learning techniques. We are aware that in this article we are mainly raising challenges without offering detailed solutions. These are categorized into three steps as presented in the architectural sketch. 
We are convinced that our research will help to overcome the mentioned problems and that we may contribute to active safety of VRUs enabling highly automated driving.

\section*{\large Acknowledgment}

This preliminary work partly results from the project DeCoInt$^2$, supported by the German Research Foundation (DFG) within the priority program SPP 1835: "Kooperativ interagierende Automobile", grant numbers DO~1186/1-1 and FU~1005/1-1 and SI~674/11-1.
The work is also supported by "Zentrum Digitalisierung Bayern".

\bibliographystyle{splncs03}

\bibliography{bibliography}

\end{document}